\documentclass[letterpaper, 10pt, conference]{ieeeconf}

\IEEEoverridecommandlockouts
\let\labelindent\relax

\usepackage[utf8]{inputenc}
\usepackage{graphics} 
\usepackage{epsfig} 
\usepackage[nolist,nohyperlinks]{acronym}
\usepackage{color}
\usepackage[ruled, vlined, commentsnumbered, linesnumbered]{algorithm2e}
\usepackage{amsmath}
\usepackage{caption}
\usepackage{subcaption}
\usepackage[font=small]{caption}
\usepackage{listings}
\usepackage{multirow}
\usepackage{enumitem}
\usepackage{siunitx}
\usepackage{graphicx}
\usepackage[dvipsnames]{xcolor}
\usepackage{tcolorbox}
\usepackage{hyperref}
\usepackage{array}
\usepackage{tabularx}
\usepackage{booktabs}
\usepackage{comment}

\definecolor{my_darkblue}{rgb}{0.239, 0.353, 0.502}
\definecolor{my_middleblue}{rgb}{0.596, 0.757, 0.851}
\definecolor{my_lightblue}{rgb}{0.878, 0.984, 0.988}
\definecolor{my_red}{rgb}{0.933, 0.4235, 0.302}

\newtcbox{\mybox}[1][red]{on line, colframe=black,colback=#1,boxrule=0.5pt,arc=4pt,boxsep=0pt,left=3pt,right=3pt,top=2pt,bottom=2pt}

\title{\LARGE \bf Automatic Extension of a Symbolic Mobile Manipulation Skill Set}

\author{Julian F\"orster$^1$, Lionel Ott$^1$, Juan Nieto$^2$, Roland Siegwart$^1$ and Jen Jen Chung$^1$%
\thanks{This work was supported in part by ABB Corporate Research and the ETH Foundation with an unrestricted gift from Huawei Technologies.}%
\thanks{$^1$ Autonomous Systems Lab, ETH Z\"urich, Zurich, Switzerland, {\tt\small <firstname>.<lastname>@mavt.ethz.ch}.}%
\thanks{$^2$ Microsoft Mixed Reality and AI Labs, Zurich, Switzerland, {\tt\small juannieto@microsoft.com}}
}

\begin{document}

\begin{acronym}
\acro{pddl}[PDDL]{Problem Domain Definition Language}
\acro{mms}[MMS]{mobile manipulation system}
\acro{mcts}[MCTS]{Monte Carlo Tree Search}
\end{acronym}

\maketitle
\thispagestyle{empty}
\pagestyle{empty}

\begin{abstract}


Symbolic planning can provide an intuitive interface for non-expert users to operate autonomous robots by abstracting away much of the low-level programming. However, symbolic planners assume that the initially provided abstract domain and problem descriptions are closed and complete. This means that they are fundamentally unable to adapt to changes in the environment or task that are not captured by the initial description. We propose a method that allows an agent to automatically extend its skill set, and thus the abstract description, upon encountering such a situation. We introduce strategies for generalizing from previous experience, completing sequences of key actions and discovering preconditions to ensure the efficiency of our skill sequence exploration scheme. The resulting system is evaluated in simulation on object rearrangement tasks. Compared to a Monte Carlo Tree Search baseline, our strategies for efficient search have on average a 29\% higher success rate at a 68\% faster runtime.

\end{abstract}

\section{Introduction}\label{sec:intro}

Today, mobile manipulators are being developed for work in unstructured \emph{human-centered} environments, i.e. spaces that are not specifically designed for deploying robots. To reach a level of ubiquity similar to their stationary counterparts in structured (e.g. manufacturing) domains, we need to shift to more flexible robot planning and execution that can adapt to changing environments and tasks. The goal is to deploy mobile manipulators in open world environments where it is impossible to foresee at design time all possible variations of tasks, disturbances, types and instances of objects, etc. In addition, a user would ideally only need to interact with the system at a mission level \cite{ajaykumar_survey_2021}, specifying high-level goals (e.g. tidy the table) for which the robot would autonomously generate a plan that it then executes.

Symbolic planning provides a framework that is conducive to this type of user interaction~\cite{long2002progress}. Here the goal is to develop domain-independent planners that, given an abstract description of an agent's skills, the entities in its environment, and an initial and desired goal state, find a sequence and parameterization of skills that achieve the goal state. By operating on an abstract level, planning for long-horizon tasks becomes tractable.

\begin{figure}
\centering
\includegraphics[width=\linewidth]{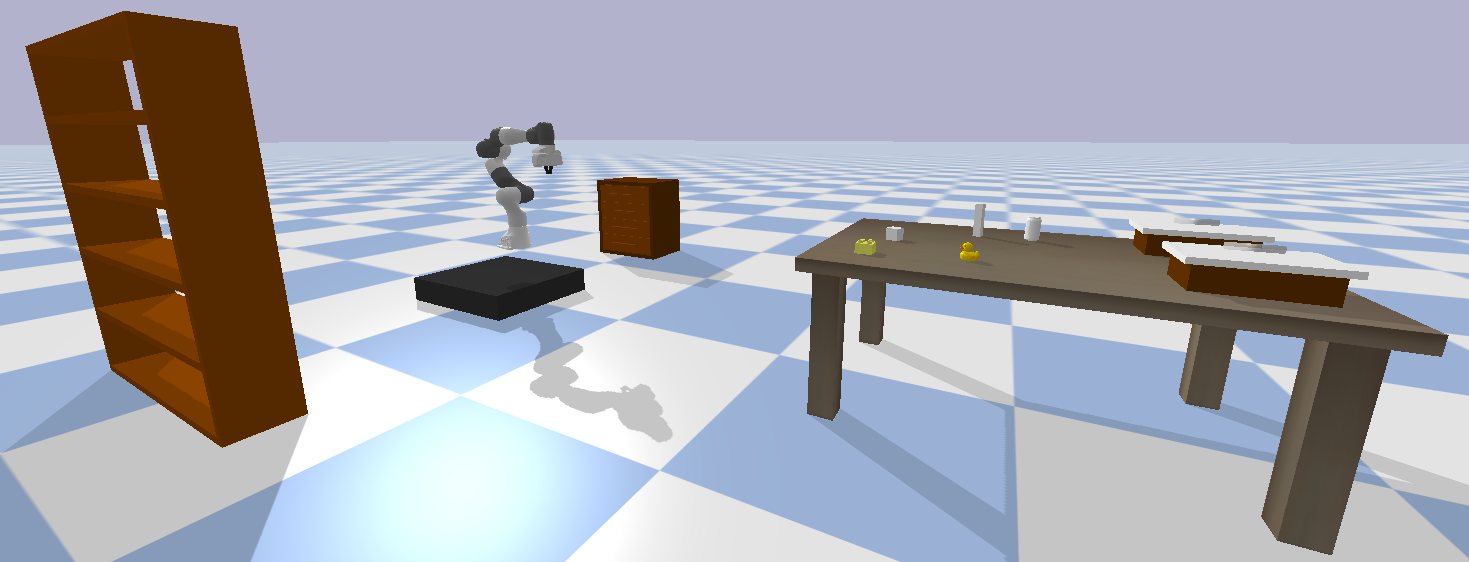}
\caption{Simulation environment setup that is used to demonstrate the proposed automatic skill set extension method. The unstructured environment is designed for object rearrangement tasks.}\label{img:simsetup}
\end{figure}

While symbolic planning can be applied very naturally to robotics problems \cite{Karpas2019}, it still relies on a complete definition of the problem at planning time. Upon encountering new tasks or new situations that cannot be captured correctly by the existing symbolic abstraction, planning will fail. In existing approaches, the symbolic abstraction would have to be updated manually to incorporate any new skills, parameters or state abstractions required to solve the new task.

This paper proposes an approach that overcomes the limitations imposed by a closed and complete world assumption. Upon encountering a failure during planning or plan execution, our method explores promising extensions to the currently defined set of skills to reach the goal.
To demonstrate our approach, we begin with a set of four basic robot skills, \textit{navigate}, \textit{grasp}, \textit{place} and \textit{move}. During exploration, sequences of these skills and suitable parameterizations (e.g. target objects or positions) are sampled and subsequently tested in a physics simulation (see Fig.~\ref{img:simsetup}). When a successful sequence is found, the symbolic description is updated to incorporate the newly gained experience, allowing the symbolic planner to solve similar tasks in the future.

Our choice of basic skills is purposefully simple to demonstrate how our approach can build up to more complex task executions. Nevertheless, even in these simple scenarios, we show that exploration can be expensive, especially when done na\"ively. Thus, we introduce the following components:
\begin{itemize}
    \item Exploration warm start based on goal entities;
    \item Generalization of new skills to other object types;
    \item Key actions for fast sequence completion; and
    \item Precondition discovery to keep the model compact.
\end{itemize}
Together, these allow the planner to efficiently discover and automatically integrate new skills to achieve increasingly complex tasks. Additionally, our framework can incorporate user demonstrations as exploration samples, accelerating the discovery of new skills.

We evaluate our approach with three different exploration strategies and compare the exploration and planning efficiency against \ac{mcts}. Our results show that the proposed algorithm leads to better performance in planning for unseen tasks, both in terms of success rate and task completion time.

\section{Related Work}\label{sec:related}

Symbolic planning lends itself very well to high-level planning in robotics \cite{Karpas2019}. In various applications \cite{Leidner2014, Srinivasa2010} impressive behaviors were achieved, using it to decide what action to take next. Typically however, the symbolic domain and problem descriptions are manually engineered, making it necessary to adapt them in case new tasks arise or if the system is to be deployed in a new environment.

Task and motion planning approaches were developed for tasks that also require careful planning on a geometric level \cite{garrett_integrated_2021, ren_extended_2021, Kaelbling2011}. 
These can deal with the complex interplay between discrete and continuous state spaces. However, their deployment to a certain class of mobile manipulation problems requires domain and task adaptations.

Garrett et al.~\cite{Garrett2018c} worked on extending the \ac{pddl} to make it more expressive and useful in a robotics context. By introducing semantic attachments, functions computed by external modules such as motion planning, inverse kinematics or sampling can be integrated naturally in the planning process. In a further extension \cite{Garrett2020}, the authors introduced planning over probability distributions which model beliefs of object states. While these approaches tackle important problems, they also require manual engineering in case new tasks arise.

Approaches exist that introduce an action hierarchy, the \textit{options} framework \cite{sutton_between_1999, kroemer_review_2021} being a prominent example. 
Bagaria and Konidaris \cite{bagaria_option_2020} propose an algorithm for discovering new options given a goal state classifier. While this idea is similar to our approach, we opted for \ac{pddl} as a representation for skills due to its human-readability which simplifies retracing decisions.
The goal of hierarchical planning as proposed by Morere et al. \cite{Morere2019a} is to combine primitive skills into meta-skills. However, it does not discover previously unknown skill effects that help to achieve unseen goals.

Another line of work aims at bridging symbolic planning and robotics applications by abstracting high-dimensional sensor data \cite{Konidaris2018, Ugur2015a, Geib2006}. The robot's actions are applied to the environment and observations are collected. Subsequently, classifiers are fit to the observations to model preconditions and effects of actions. While these methods have achieved great successes in autonomously building a symbolic description of a domain, they do not solve the problem of combining actions to achieve unseen tasks. 
Furthermore, the definition of the symbolic description based on high-dimensional sensor data makes it hard to generalize the groundings to new environments. Although progress has been made in this direction \cite{james_learning_2020}, retraining to ground the general abstractions to the scene at hand is still required.

To our knowledge, only few approaches exist for combining skills to reach new goals in a symbolic planning context. Angelov et al. \cite{Angelov2020} propose a method using the dynamics of each skill (learned or modeled) as well as a goal-scoring metric that is learned during a demonstration of the task at hand. This approach is successful at combining the robot skills at runtime guided by the goal-score metric to achieve the demonstrated goal. Another approach proposed by Strudel et al. \cite{Strudel2019} uses behavior cloning to transfer expert-defined skills from simulation to the real world and reinforcement learning to combine them. However, since both approaches are based on high-dimensional sensor data, retraining is required each time the task or the scene change.

In contrast, the approach proposed in this paper aims to learn how to solve new tasks automatically, by combining the available basic skills. By creating a modular setup and relying on existing perception tools, our approach has the potential to generalize over classes of objects as well as different scenes. Furthermore, our approach is orthogonal to the task and motion planning and hierarchical planning methods mentioned above, allowing our method to automatically extend the symbolic descriptions involved in each of them.

\section{Definitions and Problem Formulation}\label{sec:problem}

In symbolic planning, problems are typically modeled as transition systems, which consist of a set of admissible states $S$, a set of legal transitions between the states $\mathcal{T} \subseteq S \times S$, an initial state $s_0 \in S$, and a set of desired goal states $G \subseteq S$. The aim of planners is to produce a sequence of transitions leading from the initial state to a goal state.

To simplify the modeling of planning problems and to facilitate comparing symbolic planners, the Planning Domain Definition Language (PDDL) \cite{Fox2003} was created. It splits the problem into a problem-invariant \textit{domain description} and a problem-specific \textit{problem description}. The domain description consists of:

\begin{itemize}
    \item The set of \textbf{object types} $T$ organized in a hierarchy.
    \item The set of \textbf{predicates} $P$ that are available to model a state. Each predicate can accept parameters, where the type admitted for each parameter is predefined. A grounded predicate takes a binary value, indicating whether the represented property holds.
    \item The set of \textbf{actions} $A$ the agent can perform, each accepting parameters. An action's preconditions encode which predicates need to hold before the action can be executed and its effects model what predicate changes executing the action causes.
\end{itemize}

\noindent 
The problem description that needs to be updated for every new problem consists of:

\begin{itemize}
    \item A list of all \textbf{entities} (e.g. objects) $E$ in the world that are relevant for the planning problem, together with a specification of their type(s).
    \item The \textbf{initial state} $s_0$, represented by a list of grounded predicates that hold initially. All unlisted grounded predicates are considered to be false.
    \item The \textbf{goal state} $s_g$ consists of a list of grounded predicates, with an indication whether it should be true or false for each.
\end{itemize}

With these two descriptions, which can be summarized as $d = (T, P,A,E,s_0,s_g)$, off-the-shelf symbolic planners can produce a plan that solves the problem specified by the problem description only if the domain description is rich enough for a solution to exist.

We consider the problem of automatically augmenting an existing domain description that is not rich enough to model the laws governing an environment, or the capabilities of the agent. This situation can be encountered when:

\begin{description}
    \item[(C1)] The agent is tasked with achieving a newly introduced predicate (e.g. through demonstration of a goal state by a user), for which it is not modeled how to achieve it using the available actions. Characteristic of this case is that the planner fails to output a valid plan given $d$.
    \item[(C2)] The preconditions of an action performed by the agent are not fully modeled (e.g. because of a new disturbance or the agent being deployed in a new environment). This case can be detected as a failure during plan execution.
\end{description}

In the following, we split the set of actions into two subsets, $A_b$ and $A_m$ with $A=A_b \cup A_m$. Here, $A_b$ is a set of basic actions $a_b \in A_b$ available to the agent, where each action,
\begin{equation*}
    a_b = \left(\theta_{a_b}, \text{pre}(\theta_{a_b}), \text{eff}(\theta_{a_b}), \phi_{a_b}\right),
\end{equation*}

\noindent takes parameters $\theta_{a_b}$, has preconditions and effects, as well as a low-level implementation $\phi_{a_b}$ that grounds it. Meta-actions $a_m \in A_m$ are defined similarly as,
\begin{equation*}
    a_m = \left(\theta_{a_m}, \text{pre}(\theta_{a_m}), \text{eff}(\theta_{a_m}), \rho_{a_m} \right).
\end{equation*}

\noindent Instead of using a grounding $\phi$, these actions are defined using a sequence of basic actions $\rho_{a_m} = (a_{b,1},a_{b,2}, ...)$ where $a_{b,i} \in A_b$. When choosing a meta-action $a_m$, its sequence is executed under the hood. Introducing meta-actions allows the inclusion of additional preconditions and effects that are not captured by the underlying basic actions. 

For case (C1), we aim to find a (set of) new meta-action(s) $a^*_m$ such that the planner can use $d^* = (T,P,A \cup a^*_m, E, s_0, s_g)$ to devise a plan that, when executed by the agent, leads to successfully achieving $s_g$.

For case (C2), in which the plan execution fails, we aim to adapt applicable types, preconditions and effects of existing meta-actions in $A_m$, leading to $A^*_m$. If that is insufficient, we will add new meta-action(s) $a^*_m$ as necessary. As before, the planner should be able to use $d^* = (T,P,A_b \cup A^*_m \cup a^*_m, E, s_0, s_g)$ to devise a plan that, when executed by the agent, leads to successfully achieving $s_g$.

\section{Exploration for Skill Set Extension}

\subsection{Overview}

\begin{figure}
\vspace{0.2cm}
\centering
\includegraphics[width=0.8\linewidth]{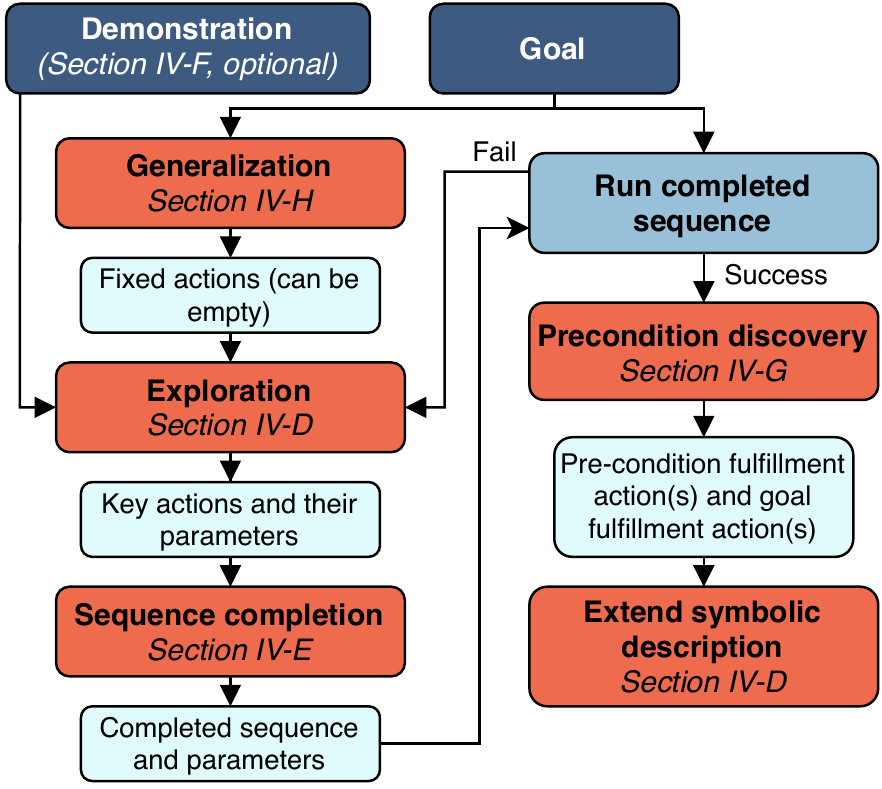}
\caption{Overview of the exploration for skill set extension with \mybox[my_darkblue]{\textcolor{white}{inputs}}, \mybox[my_red]{proposed algorithm components}, a \mybox[my_middleblue]{physics simulator} and \mybox[my_lightblue]{intermediate results}.}\label{img:expoverview}
\end{figure}

To get a better intuition of the problem and an overview of our proposed method, consider the following example. A \ac{mms} is equipped with the basic skills mentioned above (navigate, grasp, place, move) and a generic symbolic description modeling each skill's preconditions and effects as actions. Confronted with a new user-provided goal to place a cup \texttt{inside} a drawer (where \texttt{inside} is a predicate), a symbolic planner using the initial symbolic description fails to output a valid plan, since it is unknown how to reach the effect \texttt{inside} (see case (C1) in Section~\ref{sec:problem}).
From here, our proposed exploration algorithm (visualized in Fig.~\ref{img:expoverview}) takes over. The \textit{exploration} module (Section~\ref{sec:explore}) forms the core, sampling and executing sequences of basic actions and testing if the goal is achieved. To keep the sampling effective despite the large search space that is caused by long sequences and continuous parameters, we rely on \textit{sequence completion} (Section~\ref{sec:completion}). To leverage intuitions from a user, there is the option to include simple \textit{user demonstrations} (Section~\ref{sec:demonstration}). Finally, when a successful sequence is found, we use a \textit{precondition discovery} procedure (Section~\ref{sec:preconditions}) to determine which actions are required under different conditions, thus only executing the necessary actions. After this, the symbolic description is extended so that the symbolic planner can devise successful plans for this situation in the future.

Continuing our example, assume that the system is subsequently given a similar goal, e.g. to place a bottle \texttt{inside} a different drawer. Although we cannot assume that the same plan we found before will work (e.g. due to different dimensions of the involved object or drawer), we would like to benefit from previous experience. The \textit{generalization} module (Section~\ref{sec:generalization}) is responsible for that.
The following sections cover each of the mentioned components in detail.

\subsection{Parameterized Robot Skills}\label{sec:skills}

To make the high-level planning efficient enough for solving long-horizon tasks, we introduce a set of basic skills to absorb complexity. Fine-grained decisions, for example about exact trajectories to take or motor commands to issue, are handled by the skills, so that the high-level planning can focus on the goal-reaching sequence. However, as a way to influence the decisions taken by the skills, certain aspects are exposed to the high-level planner as parameters. The appropriate parameterization is part of the exploration.

A further advantage of this modular approach built on atomic skills is that the resulting system can benefit from the state of the art in specialized robot capabilities such as grasping or navigation. While we rely on four basic skills in this paper, there is the flexibility to include new skills in the future as appropriate for the application domain. 

Inspired by the atomic actions humans use to perform tasks in our environment, we selected the following four basic skills. The \textit{navigation} skill takes a goal location for the \ac{mms} as input and reaches the target while avoiding obstacles along the way. The \textit{grasp} skill, taking a goal object within reach of the robot as input, computes a suitable grasp and lets the robot arm execute it. The \textit{place} skill places an object in the robot's gripper at a given location. Finally, the \textit{move} skill is designed for force-sensitive interaction with the environment, for example, pulling open a drawer.


\subsection{Predicates}\label{sec:predicates}

Predicates that are used to define the goal need to be grounded to the environment, such that sensors can be used to judge whether a predicate currently holds. For example, consider the predicate \texttt{inside(?container, ?contained)}. For an instantiation of the parameters, such as \texttt{inside(cupboard, glass)}, a grounding function is needed that reports the state of this predicate. This can be achieved in different ways, for example with a manually implemented function making a decision based on sensor data, or a data-driven classifier.
This work focuses on modifying and extending the action set within the symbolic description. Since methods on learning predicate groundings exist (see Section~\ref{sec:related}) and can be combined with our method in the future, we assume for now that these groundings are given.


\subsection{Exploration}\label{sec:explore}

Exploration begins once (C1) or (C2) introduced in Section~\ref{sec:problem} occurs.
The aim is to find a sequence that achieves the given goal and, upon success, to extend the symbolic description appropriately. Either by adding new meta-action(s) $a^*_m$ or modifying existing ones, such that the symbolic planner is able to output valid plans for reaching the goal in the future.

To achieve this, Algorithm~\ref{alg:expl} is employed. At its core, sequences and their parameterizations are sampled, executed and tested for success. Before sampling a sequence, a sequence length needs to be determined. For this, we have three interchangeable strategies:

\begin{description}
	\item[(s1)] \textit{Alternating sequence length}. In every iteration, identified by the iteration count $n$, a different length is selected. Over the iterations, the maximum sequence lengths are $l = 1, 2, 3, ..., l_\text{max}, 1, 2, ...$. It leads to each sequence length being tested roughly the same number of times within the time budget.
	\item[(s2)] \textit{Increasing sequence length}. Each sequence length is tested for a fixed duration $t_\text{seq} = T_\text{max} / l_\text{max}$, i.e. $l = 1, 1, ..., 1, 2, 2, ..., l_\text{max}, l_\text{max}$.
	\item[(s3)] \textit{Maximum sequence length}. Every sampled sequence has the maximum length, i.e. $l = l_\text{max}, l_\text{max}, ..., l_\text{max}$.
\end{description}

\begin{algorithm}[t]
	\SetKwData{Left}{left}\SetKwData{This}{this}\SetKwData{Up}{up}
	\SetKwFunction{SamSeq}{SampleSequence}\SetKwFunction{RelObj}{FindRelevantObjects}\SetKwFunction{ExtPDDL}{ExtendSymbolicDescription}
	\SetKwFunction{SeqComp}{SequenceCompletion}\SetKwFunction{PreDisc}{PreconditionDiscovery}
	\SetKwFunction{SamParam}{SampleParameters}\SetKwFunction{RunSeq}{Execute}\SetKwFunction{TestG}{TestGoals}
	\SetKwFunction{ChooseSeqLen}{GetSequenceLength}\SetKwFunction{GetTime}{GetCurrentTime}\SetKwFunction{SeqRef}{SequenceRefinement}
 	\SetKwInOut{Input}{Input}
  
 	\Input{Set of goal states $G$, initial state $s_0$, time budget $T_\text{max}$, maximum sequence length $l_\text{max}$}
	\BlankLine
	$O\leftarrow$ \RelObj{$G$}\;
	$n\leftarrow 0$\;
	$t_\text{start} \leftarrow \GetTime{}$\;
	\While{$\GetTime{} - t_\text{start} < T_\text{max}$}{
		$l\leftarrow$ \ChooseSeqLen{$l_\text{max}$, $t$, $n$}\;
		$\tilde{S}\leftarrow$ \SamSeq{$l$}\;
			$\tilde{P}\leftarrow$ \SamParam{$\tilde{S}$, $O$}\;
			$\hat{S}, \hat{P} \leftarrow$ \SeqComp{$\tilde{S}, \tilde{P}, O, s_0$}\;
			$\text{success} \leftarrow$ \RunSeq{$\hat{S}, \hat{P}$}\;
			\lIf{\ProcNameSty{$\text{success}$}}{break}
			$n\leftarrow n+1$\;
	}
	\If{\ProcNameSty{$\text{success}$}}{
		$S, P \leftarrow$ \SeqRef{$\hat{S}, \hat{P}, G$}\;
		$C\leftarrow$ \PreDisc{$S,P,O$}\;
		\ExtPDDL{$G$,$S$,$P$,$C$}\;
	}
	\caption{Exploration}\label{alg:expl}
\end{algorithm}

 To obtain a sequence, skills are sampled uniformly from the available ones using the function \texttt{SampleSequence}. When parameterizing the sampled sequence using the function \texttt{SampleParameters}, we focus on sampling from entities that are likely to play a role in fulfilling the goal. More specifically, we sample from entities that occur in the goal specification (referred to as \textit{goal entities}) and from entities that are spatially close to a goal entity (both returned by the function \texttt{FindRelevantObjects}).

Once sequence and parameterization are determined, they are tested in a physics simulator using the \texttt{Execute} function. After every action execution, it is checked whether the goal predicates are satisfied.

When a successful sequence is found, the exploration loop stops and the symbolic domain description is extended. In preparation for this, we run \texttt{SequenceRefinement} to reduce the sequence to the minimum length that is still successful in achieving the goal by removing actions iteratively and testing the resulting sequence. Finally, the shortest sequence that is still successful is passed on.

\begin{figure}[t]
\vspace{0.15cm}
\centering
    \begin{subfigure}[b]{0.35\columnwidth}
        \centering
        \includegraphics[width=\textwidth]{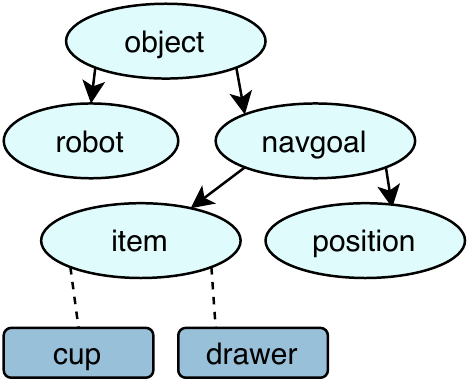}
        \caption{Before}
    \end{subfigure}
    \hfill
    \begin{subfigure}[b]{0.63\columnwidth}
        \centering
        \includegraphics[width=\textwidth]{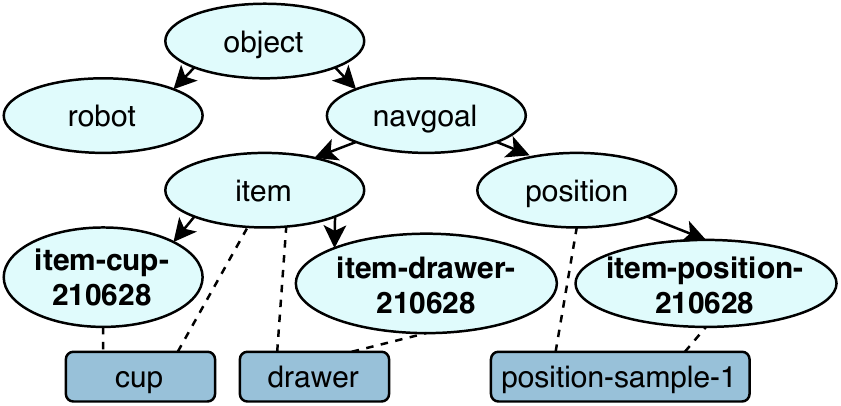}
        \caption{After}
    \end{subfigure}
    \caption{\mybox[my_lightblue]{\textcolor{black}{Types}} and \mybox[my_middleblue]{entities} assigned to them, before and after extending the symbolic description adding new sub-types (\textbf{bold}) for the cube, the cupboard and a newly introduced position sample.}\label{fig:types}
\end{figure}

For the extension of the symbolic description, implemented in \texttt{ExtendSymbolicDescription}, the collective preconditions, parameters and effects of all actions in the sequence are determined from existing abstract descriptions and from observations of running the sequence.
Furthermore, we want to avoid that our symbolic description allows applying meta-actions to entities that are incompatible in reality. For this, new symbolic types are introduced for all parameter variables, branching off the original types of the entities assigned to the parameter variables. These entities are then assigned the new types in addition to their existing types. For an example, refer to Fig.~\ref{fig:types}. New types can later be assigned to other entities once it has been verified that they are compatible with the new action (see Section~\ref{sec:generalization}).


\subsection{Sequence Completion}\label{sec:completion}

Output sequences for mobile manipulation tasks can be fairly long. However, we note that often just a small subset of \emph{key actions} are crucial for the success of the task, while the other actions merely fulfill preconditions of the key actions. Based on this insight, we can drastically reduce our exploration effort by sampling short sequences on line 6 of Algorithm~\ref{alg:expl} and leveraging the symbolic planner to solve for the corresponding complete and feasible sequence in our \texttt{SequenceCompletion} routine. We refer to the actions in the uncompleted short sequence as \textit{key actions}.

In terms of our running example, say that we want to reach the goal ``cup in drawer". A successful sequence would be:

\begin{lstlisting}
["navigate to drawer","grasp drawer",
 "move drawer","place drawer",
 "navigate to cup","grasp cup",
 "navigate to drawer","place cup"].
\end{lstlisting}

\noindent With sequence completion, we can infer this sequence from the considerably shorter sequence:

\begin{lstlisting}
["move drawer","place cup"].
\end{lstlisting}

\noindent Note that remaining unspecified parameters still need to be discovered through exploration given initial and goal states.

Consequently, the task of finding a sequence that achieves a certain goal is reduced to finding the key actions from which such a sequence can be constructed using the symbolic planner, thus greatly reducing the search space.

Our procedure of sequence completion is laid out in Algorithm~\ref{alg:compl}. 
For each key action $\tilde{a}_i$ in the sampled sub-sequence, we solve a symbolic planning problem (using \texttt{SolvePDDL}) that has the preconditions of the current key action as desired goal, resulting in a fill sequence $\hat{S}_i = ({}_i \hat{a}_1, ..., {}_i \hat{a}_{\hat{n}_i})$ and corresponding parameterizations $\hat{P}_i$. Both are added to the completed sequence $\hat{S}$ and parameters $\hat{P}$. Furthermore, before the next iteration, effects of both the fill sequence and the currently considered key action are applied to the state (using \texttt{ApplyEffects}), which was initialized with the initial state $s_0$.

\begin{algorithm}
	\SetKwProg{Fn}{Function}{}{end}
	\SetKwFunction{ApplyEffects}{ApplyEffects}\SetKwFunction{GetParam}{GetParameters}
	\SetKwFunction{GetPre}{GetPreconditions}\SetKwFunction{Solve}{SolvePDDL}
 	\SetKwInOut{Input}{Input}\SetKwInOut{Output}{Output}
 	\SetKwData{Action}{action}
 	  
 	\Input{Sub-sequence $\tilde{S}$, parameters $\tilde{P}$, set of relevant entities $O$, initial state $s_0$}
	\Output{Completed sequence $\hat{S}$, parameters $\hat{P}$, indices of key actions $I_\text{key}$}
	\BlankLine

	\Fn{\SeqComp{$\tilde{S}$, $\tilde{P}$, $O$, $s_0$}}{
		$\hat{S}, \hat{P}, I_\text{key} \leftarrow [\;], [\;], [\;]$\;
		$X \leftarrow s_0$\tcc*[r]{track current state}
		\ForEach{$\tilde{a}_i \in \tilde{S}$}{
			$p\leftarrow$ \GetPre{$\tilde{a}_i$}\;
			$\theta\leftarrow$ \GetParam{$\tilde{a}_i$, $\tilde{P}$}\;
			$\check{S}, \check{P}\leftarrow$ \Solve{initial $=X$, goal $=p$}\;
			$X\leftarrow$ \ApplyEffects{$X, \check{S}, \check{P}$}\;
			$X\leftarrow$ \ApplyEffects{$X$, $\tilde{a}_i$, $\theta$}\;
			$\hat{S}\leftarrow \hat{S} + \check{S}\; + \;\tilde{a}_i$\;
			$\hat{P}\leftarrow \hat{P} + \check{P}+\theta$\;
			$I_\text{key}.\text{append}\left(\text{length}(\hat{S}) - 1 \right)$\;
		}
		\Return{\ProcNameSty{$\hat{S}$, $\hat{P}$, $I_\text{key}$}}\;
	}
	\caption{Sequence completion}\label{alg:compl}
\end{algorithm}

After iterating over all key actions, the completed sequence returned by \texttt{SequenceCompletion} has the form,
\begin{equation*}
	\hat{S} = \Big(\underbrace{{}_1 \hat{a}_1, ..., {}_1 \hat{a}_{\hat{n}_1}}_{\hat{S}_1}, \tilde{a}_1, \underbrace{{}_2 \hat{a}_1, ..., {}_2 \hat{a}_{\hat{n}_2}}_{\hat{S}_2}, \tilde{a}_2, ..., {}_{\tilde{n}} \hat{a}_{\hat{n}_{\tilde{n}}}, \tilde{a}_{\tilde{n}}\Big).
\end{equation*}


\subsection{Demonstrations}\label{sec:demonstration}

The concept of sequence completion allows for another elegant way to make the exploration more efficient. Since humans are very good at planning for manipulation tasks, it seems natural to leverage a user's knowledge for extending the capabilities of an \ac{mms}. However, this should be possible without requiring expert knowledge. To achieve this, a user can supply our system with one or several key actions that will likely lead to successful achievement of a goal. In addition, crucial parts of the parameterization can be given. The exploration procedure with sequence completion can then be used to fill in any missing parameters of the key actions as well as any actions that are missing before or in between the key actions. All in all, this feature provides an interesting middle ground, making it easy for the user to bring in a demonstration without the need to specify all details of a sequence and at the same time drastically reducing the search space during exploration.


\subsection{Precondition Discovery}\label{sec:preconditions}

In practice it can happen that not all steps of a discovered sequence are needed every time a similar goal needs to be reached. For example, if during exploration, an obstacle was present and the agent correctly learned that the obstacle needs to be removed before the goal can be achieved, the actions to remove the obstacle will only be needed in the future if the obstacle is present in the individual situation. In our running example, the closed drawer can be seen as such an obstacle, which only needs to be opened if closed.

The symbolic description should correctly capture what parts of a sequence actually achieve the goal and what parts solely fulfill preconditions for the goal-achieving actions. We tackle this \texttt{PreconditionDiscovery} (Algorithm~\ref{alg:preconddisc}) by simulating a discovered sequence and observing any unmodelled predicate changes (\textit{side effects}) that involve goal entities and entities that are spatially close to a goal entity (returned by \texttt{FindRelevantPredicates}). Each change (detected using \texttt{MeasurePredicates} and \texttt{DetectChanges}) is considered a candidate for a precondition.

However, to avoid adding superfluous preconditions and thus fragmenting the discovered sequence more than necessary, we filter the candidates, removing:
\begin{itemize}
	\item Candidates that are contained in the goal specification,
	\item Side effects of the final action in the sequence, and
	\item Candidates that get toggled throughout the sequence execution, i.e. set and later unset or vice versa.
\end{itemize}

\begin{algorithm}[t]
	\SetKwProg{Fn}{Function}{}{end}
	\SetKwFunction{RelPred}{FindRelevantPredicates}\SetKwFunction{Measure}{MeasurePredicates}\SetKwFunction{DetectCh}{DetectChanges}
	\SetKwFunction{FiltG}{FilterGoals}\SetKwFunction{FiltT}{FilterToggling}\SetKwFunction{FiltLA}{FilterLastAction}
 	\SetKwInOut{Input}{Input}\SetKwInOut{Output}{Output}
 	\SetKwData{Action}{action}
 	  
 	\Input{Sequence $S$, parameters $P$, set of relevant objects $O$}
	\Output{Precondition candidates $C$}
	\BlankLine

	\Fn{\PreDisc{$S$, $P$, $O$}}{
		$C\leftarrow [\;]$\tcp*[r]{precondition candidates}
		$\rho \leftarrow$\RelPred{$O$}\;
		$p_\text{post} \leftarrow$\Measure{$\rho$}\;
		\For{$a \in S$}{
			$p_\text{pre} \leftarrow p_\text{post}$\;
			$\theta\leftarrow$ \GetParam{$a$, $P$}\;
			\RunSeq{$a$, $\theta$}\;
			$p_\text{post} \leftarrow$\Measure{$\rho$}\;
			$p_\text{new}\leftarrow$ \DetectCh{$p_\text{pre}, p_\text{post}$, $a$, $\theta$}\;
			$C\leftarrow C + p_\text{new}$\;
		}
		\tcp{Filter candidates}
		$C\leftarrow$\FiltG{$C$}\;
		$C\leftarrow$\FiltLA{$C$}\;
		$C\leftarrow$\FiltT{$C$}\;
		\Return{\ProcNameSty{$C$}}\;
	}
	\caption{Precondition discovery}\label{alg:preconddisc}
\end{algorithm}
\vspace{-0.25cm}

\subsection{Generalization and Reuse of Previous Experience}\label{sec:generalization}

It is common in mobile manipulation applications that over time, similar goals need to be achieved, but for different objects and circumstances. In such a case, we want our system to leverage previous experience in order to find a solution without exploration from scratch. Assume that in our running example, the agent already knows how to place the cup inside the drawer and now wants to place the plate inside the drawer. Ideally, the existing experience should be used when figuring out how to achieve the new goal.

In this work, we achieve this by first relaxing the type specifications of entities that are part of the goal specification, allowing them to generalize to all available types and thus fit any parameter of any action. If the symbolic planner succeeds in finding a plan under these conditions, the actions forming that plan may help to achieve the goal. After extracting the action that actually achieves the goal (called the \textit{generalization candidate}), exploration continues as in Section~\ref{sec:explore}. However, every time \texttt{SampleSequence} is called, the generalization candidate is taken as a given part of the sampled sequence. This has the purpose of finding auxiliary actions and parameterizations that, together with the extracted action, form a successful sequence.

Once this sequence is found, the symbolic description is adapted. If the generalization candidate turns out to be part of the goal-reaching sequence, the corresponding action description and the types of the goal entities are adjusted to be compatible with one another.

Apart from making the exploration more efficient by generalizing previous experience, this procedure has the advantage that it contributes to making the action space in the symbolic description as large as necessary, but keeps it as small as possible at the same time.

\newcommand{\abbrvfirstpro}[0]{\textit{container procedure}}

\section{Experiments and Results Analysis}

\subsection{Setup}

We ran various experiments in a PyBullet physics simulation environment \cite{coumans2019} to evaluate our method. The simulation environment is shown in Fig.~\ref{img:simsetup}. All experiments were conducted using an \textit{Intel Core i7-9750H} laptop CPU.

Since this work focuses on high-level planning, we used simplified implementations for the robot skills. The navigation skill teleports the robot in simulation to the collision-free location closest to the desired goal location. For the grasping skill, grasp poses are pre-defined for all objects.
As symbolic planner, we use \textit{Metric-FF} \cite{Hoffmann2003}. Our algorithm writes symbolic description files, sends them to the planner and parses the planner's output for further processing.

\subsection{MCTS Baseline}\label{sec:baseline}

We compare our approach against \ac{mcts}. This method successively builds a tree, where nodes represent world states and edges represent actions available to the agent. To ensure comparability, the baseline shares the actions (including implementation) and the simulation for determining action outcomes with our method. Our implementation of \ac{mcts} is inspired by \cite{ren_extended_2021}. At every iteration, a decision is made on whether to expand the current node (i.e. try a new action from the node's state) or to select a child of the current node and to continue from there. The decision is based on a progressive widening law \cite{hutchison_continuous_2013} as follows. The node is expanded if $\lfloor N^\alpha \rfloor > \lfloor (N-1)^\alpha \rfloor $ is true, where $\lfloor . \rfloor$ is the floor operation, $N$ is the number of times the current node was already visited, and $\alpha \in [0,1]$ is a parameter controlling the balance between expanding and selecting a child (the higher $\alpha$, the more we expand). In this work, we use $\alpha=0.6$. When expanding, we only select from actions that are feasible in the current state (corresponds to preconditions being met). Furthermore, the sampling of parameterizations is limited to objects of interest in the same way as it is for our method, as explained in Section~\ref{sec:explore}.

\subsection{Results}\label{sec:quantres}

\renewcommand{\tabularxcolumn}[1]{m{#1}}
\begin{table}
\vspace{0.2cm}
\caption{Experiment scenarios that were used for evaluations. ``ID" stands for experiment ID, ``Seq. len." is the minimum length of a successful sequence, ``\# key act." refers to the minimum number of key actions in a successful sequence, and ``Prior" refers to the information available to the algorithm at planning time.}\label{tab:expscenarios}
\vspace{-6pt}
\begin{center}
{\setlength\tabcolsep{3.5pt}
\setlength\extrarowheight{3pt}
\begin{tabularx}{\linewidth}{
	>{\raggedright\arraybackslash}m{0.44cm}
	>{\raggedright\arraybackslash}m{1.4cm}
	>{\raggedright\arraybackslash}X
	>{\centering\arraybackslash}m{0.53cm}
	>{\centering\arraybackslash}m{0.7cm}
	>{\raggedright\arraybackslash}m{0.73cm}
}
\toprule
ID & Goal & Initial state & Seq. len & \# key act. & Prior \\
\midrule
(a) & Cube on cupboard & Cube lying on table & 4 & 1 & None \\
(b) & Tall box inside shelf & Tall box standing on table & 4 & 1 & None \\
(c1) & Cube inside container1 & Cube on table, container1 covered with lid & 8 & 2 & None \\
(c2) & Duck inside container1 & Duck on table, container1 covered with lid & 8 & 2 & After (c1) \\
(c3) & Cube inside container2 & Cube on table, container2 covered with lid & 8 & 2 & After (c1) \\
(c4) & Duck inside container2 & Duck on table, container2 covered with lid & 8 & 2 & After (c1) \\
(d1) & Cube inside container2 & Cube inside container1, both containers covered with lids & 12 & 3 & None \\
(d2) & Duck inside container2 & Duck inside container1, both containers covered with lid & 12 & 3 & After (d1) \\
\bottomrule
\end{tabularx}}
\end{center}
\end{table}

We conducted experiments with different variants of our algorithm as well as the \ac{mcts} baseline on the scenarios of increasing length and difficulty shown in Table~\ref{tab:expscenarios}. Scenarios (c2)-(c4) as well as (d2) test the benefit of prior knowledge.

First, we investigated the influence of the three sampling strategies (s1)-(s3) introduced in Section~\ref{sec:explore}. They are labelled as ``ours, alternating (OA)", ``ours, no alternating (ONA)", and ``ours, full length (OFL)". For these three variants, the \textit{navigation} and \textit{grasp} skills were excluded from key action sampling to test how the number of actions influences the performance. To compare, a variant on (s3) was included that samples from all basic skills, termed ``ours, full length, all skills (OFLA)". A fifth instance of our method, labelled ``ours, from demo. (OD)" was provided with a task-specific demonstration.
For example, we used \texttt{["place", "place"]} as the sequence demonstration and \texttt{[\{"obj": "lid1"\}, \{"obj": "cube"\}]} as the parameter demonstration for scenario (c1). For the other scenarios, analogous demonstrations were provided.
Finally, \ac{mcts} was included as a baseline. For all methods, a maximum sequence length to sample needs to be specified. With the intention to overestimate the truly needed length, we chose 4 for scenarios (a)-(c4) and 6 for scenarios (d1)-(d2) for our method. We observed that on average, there are 4 actions per key action in the expanded sequence, leading to selecting a maximum search tree depth of 16 and 24, respectively for \ac{mcts}.

Each of these six algorithms was run 20 times on each of the scenarios (Table~\ref{tab:expscenarios}), each with a time budget of 900 seconds. Before each run, the simulated scene was reset to its initial state, depicted in Fig.~\ref{img:simsetup}. The symbolic description available to the agent in the beginning of a run was also reset to the four basic skills except in cases where we were testing our method with initial knowledge resulting from a previous successful exploration.

\begin{figure*}
    \centering
    \null\hfill
	\includegraphics[height=6.0cm]{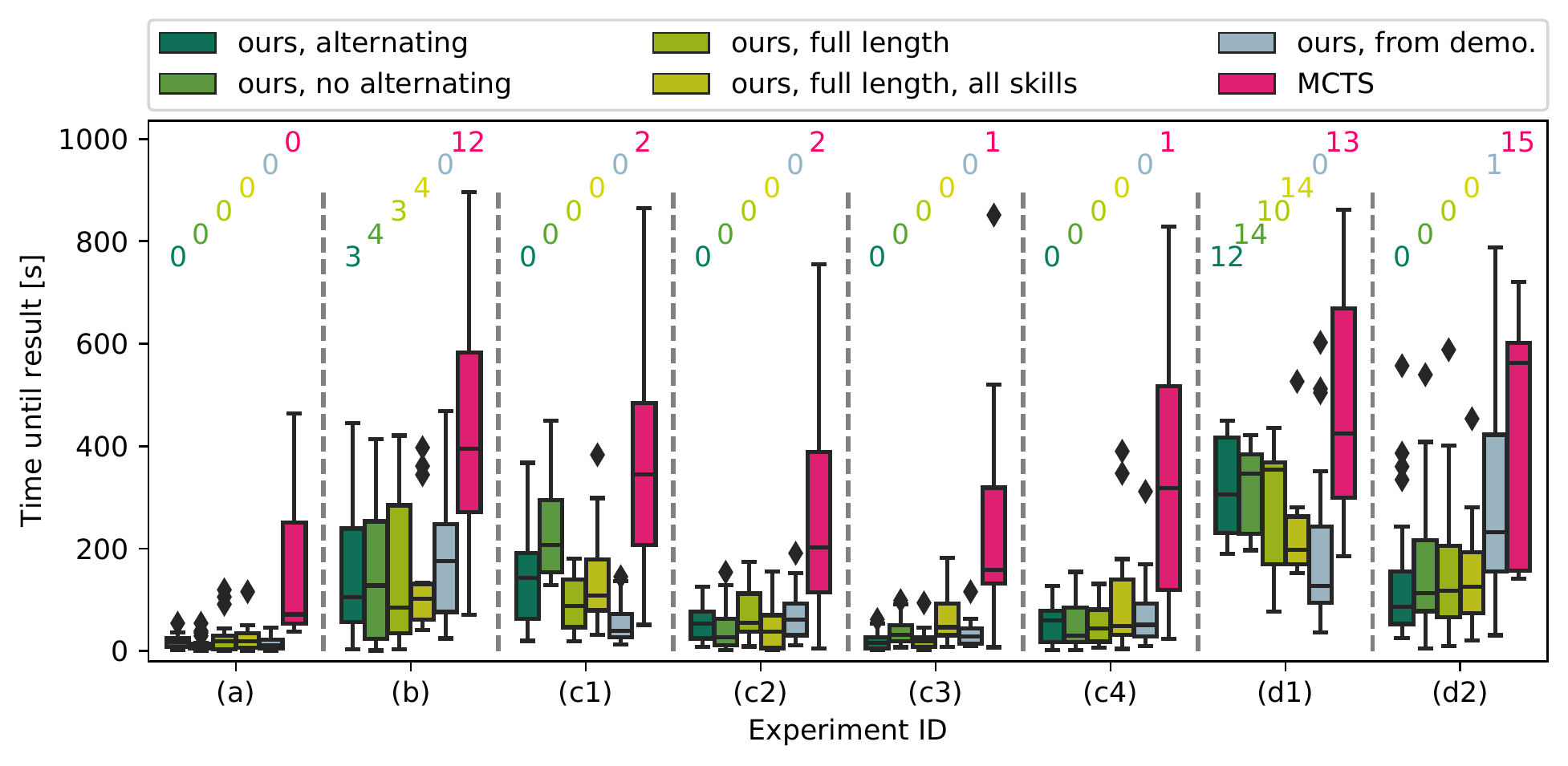}
	\hfill
	\includegraphics[height=6.0cm]{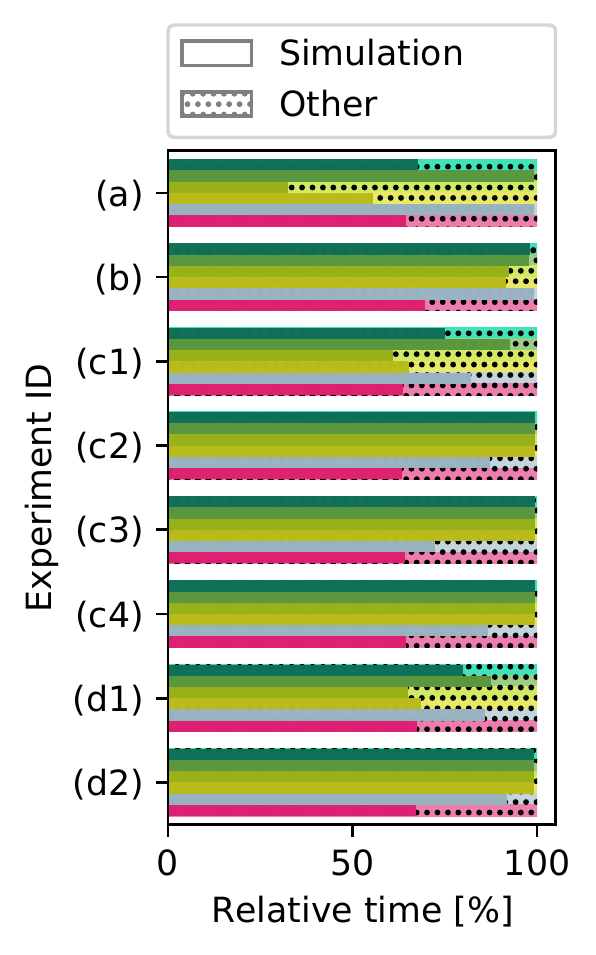}
    \hfill\null\vspace{-0.25cm}
    \caption{(left) Runtime comparison of all sequence exploration methods for the scenarios shown in Table~\ref{tab:expscenarios}. Each method was run 20 times for each scenario, with a time budget of 900 seconds. The numbers at the top of the diagram indicate how many of the 20 trials timed out before finding a feasible solution (lower is better). Only the times taken by successful runs are included in the plot. (right) Average simulation time as a percentage of the total planning time for each method and all experiment scenarios.}
    \label{img:results}
    \vspace{-0.5cm}
\end{figure*}

The results of these experiments are shown in Fig.~\ref{img:results}. In Fig.~\ref{img:results} (left), we report success rates as well as the distribution of time spent until a successful solution was found. The performance of the \ac{mcts} baseline is consistently worse in all experiments, both in success rate and execution time. We believe that the main reason for this is that our method benefits from sequence completion and can thus focus on the shorter key action sequences.
Running our method without sequence completion as an ablation to confirm this is not reasonable, since exploiting logic reasoning to generate feasible sequences is an integral component of our method.
Additionally, \ac{mcts} does not benefit from previous experiences, having to explore from scratch every time.

For all methods, the computational complexity tends to increase and the success rate tends to drop with increasing sequence length. An exception is experiment (b), where a precise placing location has to be found by all methods, which increases the average duration until success. OD has the largest success rate, benefiting from the provided demonstration. Between the methods that have no prior knowledge, we expected ONA to fare best since it starts by exclusively sampling sequences of length 1, which is exactly the minimum number of key actions needed in this case. However, the results show that longer sequence lengths during sampling are not disadvantageous. For problems where sampling longer sequences is required, ONA performs the worst as expected, because its sampling budget for shorter-length sequences must be exhausted before longer sequences are considered. For OA and OFL(A), this is not the case, resulting in a better performance in scenarios (c1) and (d1).

In scenarios where prior knowledge is available ((c2)-(c4), (d2)), performance is improved for all variants of our method, showing the benefit of generalization. Note that OA, ONA and OFL(A) perform similarly in these scenarios, since the considered sequence length is the same for all when generalizing. Further, the generalization is skipped when a user demonstration is available (i.e. for OD). Therefore, OD performs the same, independent from whether prior experience is given.

The experiments also showed that the complexity increases dramatically for longer sequence lengths. In scenario (d1), the success rates are low, except for OD, which can be expected. Apart from that, OFL also performs significantly better than the other variants. Since OFL always attempts sequences of the maximum length, it does not waste any exploration effort on shorter sequences, which in this scenario cannot solve the task.
Nevertheless, all variants again perform similarly in scenario (d2), where the number of failures can be reduced to zero through generalization.

Comparing OFLA to the other variants showed that the advantage of sampling from fewer skills is not as pronounced as initially expected. This is positive, indicating that our method can generalize to other skill sets without manually selecting skills to exclude. Nevertheless, the option to exclude skills could still be beneficial for very long scenarios. Investigating this is left to future work.

Overall, it is clearly beneficial to know the exact sequence length required to solve the task as this significantly refines the exploration effort. Of course, this information is not available at planning time in most cases. Since it is possible to extract a shorter sequence from a longer one via our precondition discovery, OFL(A) may be preferential to the other sequence exploration variants. OFL(A) is the most successful for long sequence lengths while only incurring a small planning time overhead at short sequence lengths.

\newlength{\twosubht}
\newsavebox{\twosubbox}

\section{Conclusion and Future Work}

We presented a symbolic planning system that automatically extends its abstract skill set to achieve goals for which either the symbolic planning or the plan execution failed. Our results show that the proposed algorithm does so in a consistent way, extending the symbolic description as much as necessary to achieve the goals, while keeping it as small as possible, to make sure that planning stays sound and tractable. Furthermore, our measures to run the exploration more efficiently greatly reduce the computational complexity, outperforming the \ac{mcts} baseline we compare against, thus increasing our method's value in practice.

While these results are promising, our system still has limitations. In the future, to avoid the need for manually grounding predicates, we plan to learn models of predicates based on demonstrations and interactions with the environment. Furthermore, we aim to leverage data gathered in multiple encounters of a task to increase the accuracy of preconditions and effects of action abstraction beyond what is possible in single encounters of such a task.




\bibliographystyle{IEEEtran}
\bibliography{main.bib}

\end{document}